\documentclass[conference]{IEEEtran}
\IEEEoverridecommandlockouts

\usepackage{textcomp}

\usepackage{amsmath,amssymb,amsfonts}
\usepackage{graphicx}
\usepackage{booktabs}
\usepackage{multirow}
\usepackage{microtype}
\usepackage{url}
\usepackage{hyperref}
\hypersetup{hidelinks}
\usepackage{xcolor,soul}
\usepackage{booktabs}
\usepackage{multirow}  
\usepackage{booktabs} 
\usepackage{cite}

\def\BibTeX{{\rm B\kern-.05em{\sc i\kern-.025em b}\kern-.08em
    T\kern-.1667em\lower.7ex\hbox{E}\kern-.125emX}}
\begin{document}

\title{On-Device Fine-Tuning via Backprop-Free Zeroth-Order Optimization
{}
\thanks{This project is funded by the Advanced Research + Invention Agency (ARIA). The work of O. Simeone and B. Rajendran was also supported by Open Fellowships of the EPSRC (EP/W024101/1,  EP/X011356/1) and by the EPSRC project (EP/X011852/1). The work of O. Simeone was also supported by  the European Research Council (ERC) under the European Union’s Horizon Europe Programme (grant agreement No. 101198347).}
}
\author{Prabodh Katti$^{\ddagger}$, Houssem Sifaou$^{\ddagger}$, Sangwoo Park$^{\dagger}$, Bipin Rajendran$^{\ddagger}$, and Osvaldo Simeone$^{\ddagger}$\\[0.5em]
{\small $^{\ddagger}$Institute for Intelligent Networked Systems, Northeastern University London, London E1 8PH, UK}\\
{\small $^{\dagger}$Department of Engineering, King's College London, London WC2R 2LS, UK}
}


\maketitle

\begin{abstract}
 On-device fine-tuning is a critical capability for edge AI systems, which must support adaptation to different agentic tasks under stringent memory constraints. Conventional backpropagation (BP)-based training requires storing layer activations and optimizer states, a demand that can be only partially alleviated through checkpointing. In edge deployments in which the model weights must reside entirely in device memory, this overhead severely limits the maximum model size that can be deployed. Memory-efficient zeroth-order optimization (MeZO)  alleviates this bottleneck by estimating gradients using forward evaluations alone, eliminating the need for storing intermediate activations or optimizer states. This enables significantly larger models to fit within on-chip memory, albeit at the cost of potentially longer fine-tuning wall-clock time. This paper first  provides a mathematical estimate of the relative model sizes that can be accommodated under BP and MeZO training. We then numerically validate the analysis, demonstrating that MeZO exhibits accuracy  advantages under on-device memory constraints, provided sufficient wall-clock time is available for fine-tuning.
\end{abstract}


\section{Introduction}
\label{sec:intro}

\noindent \textbf{Context and Motivation:}
A key avenue for advancing artificial intelligence (AI) performance is tailoring models to specific tasks and user needs via fine-tuning \cite{lialin2023scaling}, particularly for agentic systems \cite{chen2023fireact}. User demands for modern AI models are not static, making fine-tuning a recurring necessity rather than a one-time cost. On-device fine-tuning directly addresses this need, enabling edge devices to adapt models without relying on external servers. 

However, fine-tuning may be substantially more memory-intensive than inference. In edge deployments, the model must ideally fit in the device memory \cite{slamanig2025llms, 10.1007/978-3-031-99965-9_31}. However, in order to enable fine-tuning, one is forced to implement smaller-scale models as compared to inference-only settings in order to accommodate the memory requirements of optimization \cite{tenison2026parameter}. This paper examines a possible solution to this problem, moving away from conventional backpropagation (BP) towards  memory-efficient zeroth-order optimization (MeZO)  \cite{malladi2023fine} (see \cite{wang2026universally} for a recent review).  An experimental demonstration of MeZO on edge devices was described in \cite{peng-etal-2024-pocketllm}. 

BP incurs a heavy memory overhead because it requires storing intermediate activations for every layer, resulting in memory demands far exceeding those of inference \cite{korthikanti2023reducing, malladi2023fine,tenison2026parameter}. Even with techniques such as activation checkpointing \cite{panchal2025avoidbp}, BP often remains impractical for resource-constrained devices. For instance, fine-tuning a 13B-parameter transformer can require an order-of-magnitude more memory than a single forward pass, implying that even an 80~GB GPU can only accommodate models on the order of $3$ billion parameters using  BP \cite{malladi2023fine}.

\begin{figure}
    \centering
    \includegraphics[width=\linewidth]{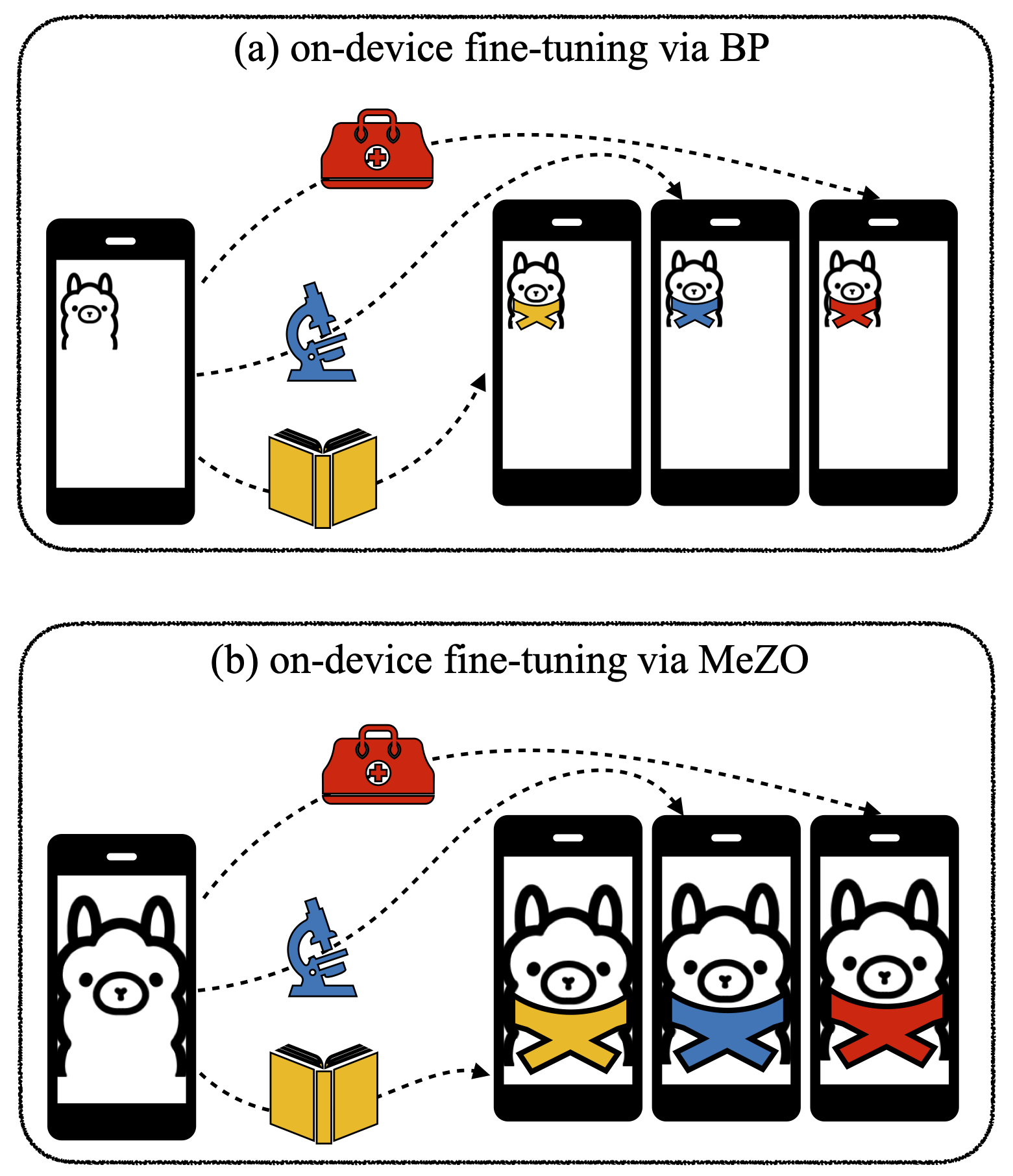}
    \caption{On-device fine-tuning: (a) backpropagation (BP)-based fine-tuning requires significantly more memory than inference (see Fig. \ref{fig:BPvsMeZO}), limiting model size on the device; while (b) MeZO-based fine-tuning \cite{malladi2023fine} only carries out inference steps, enabling deployment of significantly larger, more capable models at the edge.}
    \label{fig:placeholder}
\end{figure}



Rooted in simultaneous perturbation stochastic approximation (SPSA) \cite{spall1998overview}, MeZO approximates gradients using only forward passes. While this eliminates the need to store intermediate activations, MeZO's convergence rates degrade with the size of the parameter vector \cite{duchi2015optimal}. Recent advances, however, have shown that strong pre-training substantially reduces the effective dimensionality of the parameter space to be updated, making MeZO surprisingly competitive for fine-tuning \cite{malladi2023fine}.



Since the initial demonstration in \cite{malladi2023fine}, researchers have developed several extensions of MeZO. Notably, MeZO has been extended to incorporate parameter-efficient fine-tuning (PEFT) via sparse updates  \cite{liu2024sparse,guo2025zerothorder} or via low-rank adaptation (LoRA) \cite{malladi2023fine}. PEFT is particularly well suited for MeZO as it reduces variance and enhances convergence of fine-tuning. Other related works include \cite{yang2024adazeta} and \cite{chen2025loho}. 


\begin{figure}
    \centering
    \includegraphics[width=\linewidth]{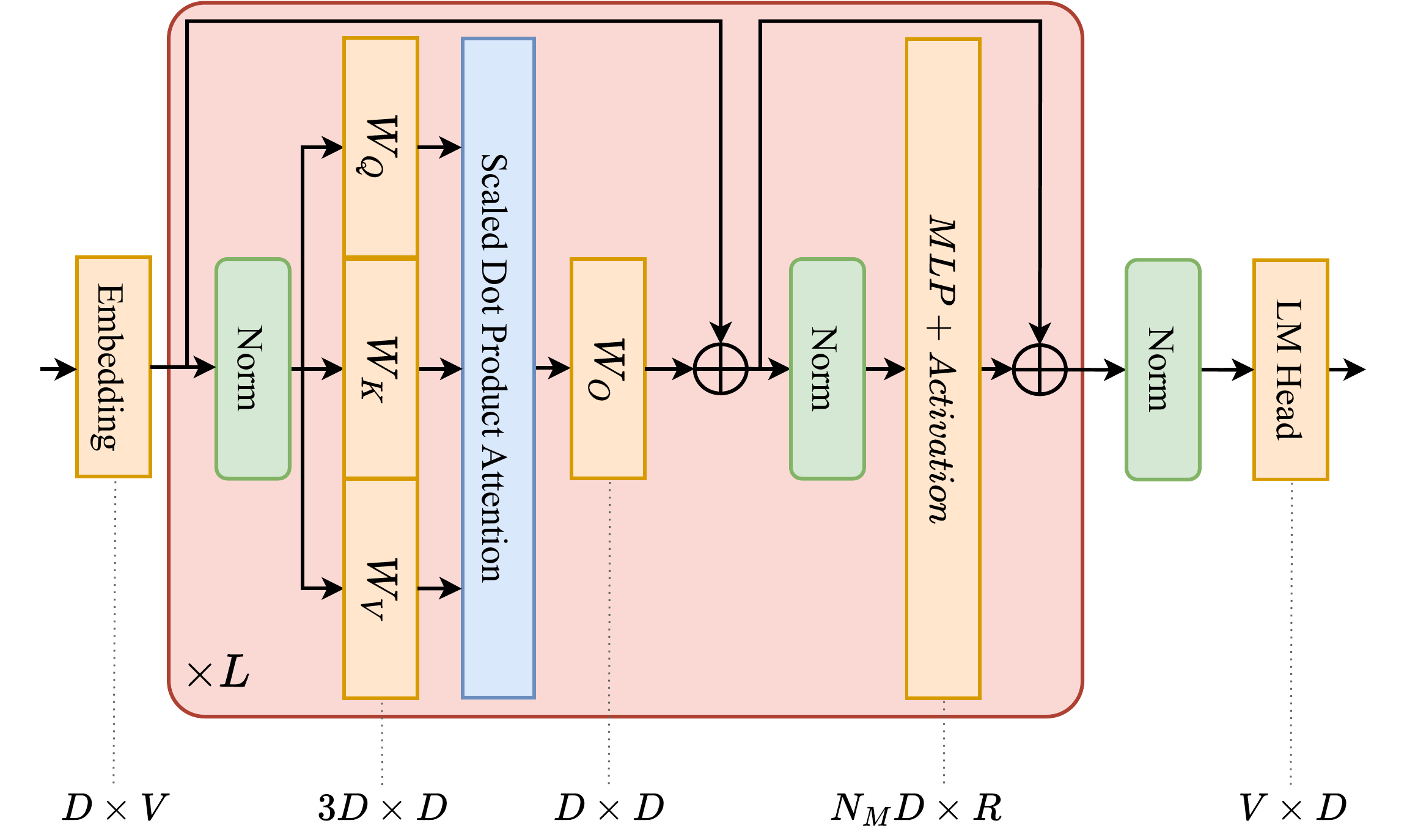}
    \caption{Illustration of a basic decoder-only Transformer model, highlighting the major contributors to parameter count. Non-parametric operations such as RoPE encoding \cite{10.1016/j.neucom.2023.127063} and parameter count for minor contributors, such as normalization layers, are not shown for clarity. }
    \label{fig:param_llama}
\end{figure}

\begin{figure*}[ht]
    \centering
    \includegraphics[width=\linewidth]{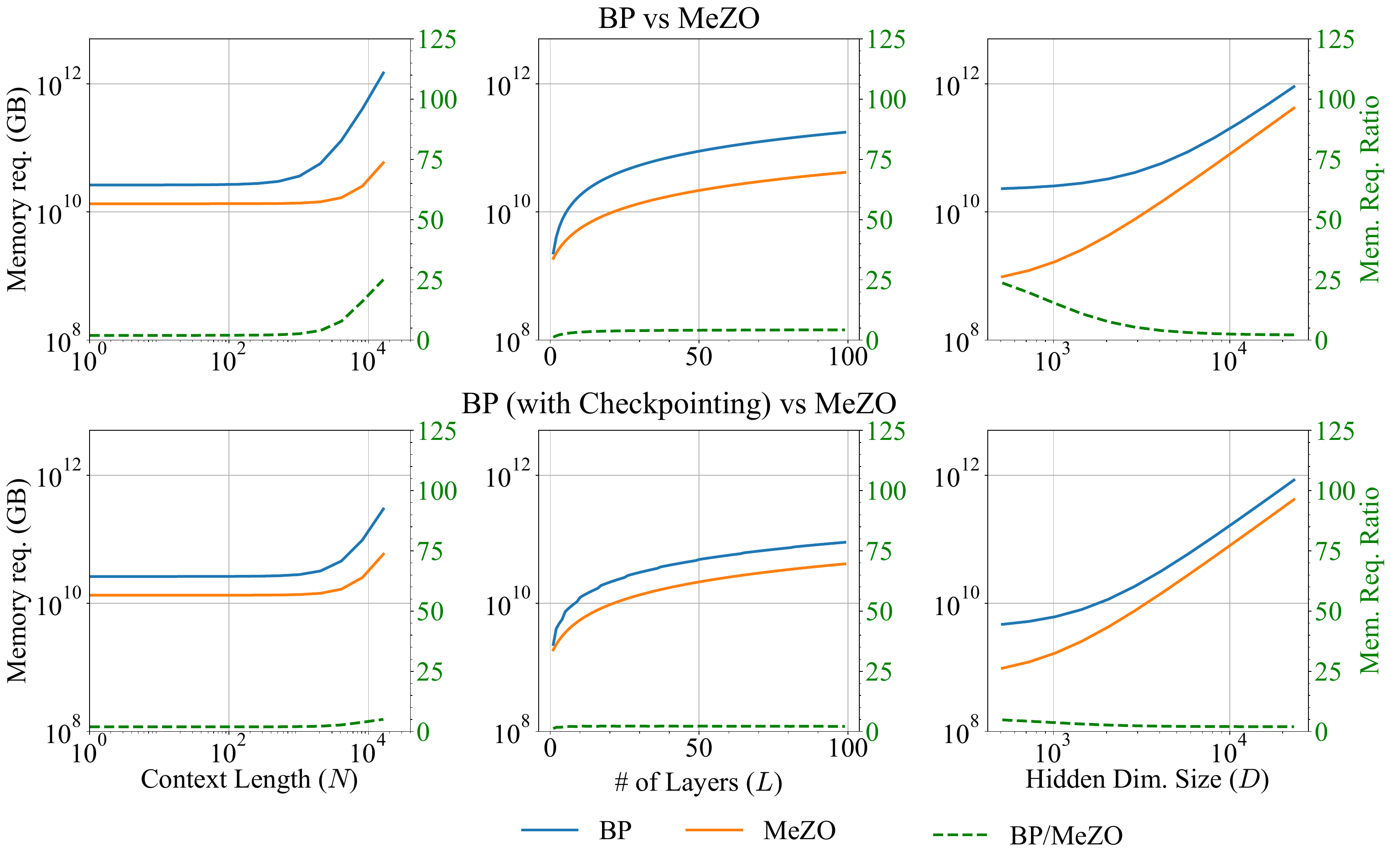}
    \caption{Memory requirements for BP- and MeZO-based fine-tuning, including conventional BP (top row) and BP with checkpointing (bottom row). The solid lines and the primary y-axis correspond to the total memory consumed by BP and MeZO using \eqref{eq:mbpsgd}, while the secondary y-axis with a green colored dashed line represents the MeZO over BP ratio of memory requirements.  }
    \label{fig:BPvsMeZO}
\end{figure*}

\noindent \textbf{Main Contributions:}
This paper provides a comprehensive study of on-device fine-tuning for LLMs, comparing BP and MeZO approaches. Our main contributions are:\\
\noindent $\bullet$ \textbf{Memory analysis of BP vs. MeZO:} We analyze model sizes that can be fine-tuned under fixed on-chip memory budgets, showing that MeZO can accommodate models at least $2 \times$ larger than BP within the same budget. This advantage grows with longer context windows, reaching up to $25 \times$ as the context length increases (see Fig.~\ref{fig:BPvsMeZO}).\\ 
\noindent $\bullet$  \textbf{Empirical evaluation:} We numerically validate our theoretical analysis, demonstrating that MeZO achieves superior accuracy under on-device memory constraints, provided sufficient wall-clock time is available for fine-tuning. We further evaluate both full-parameter and PEFT variants across multiple model scales and datasets.

\section{Memory Requirements of BP and MeZO}
\label{sec:memory}
In this section, we analyze the memory requirements of BP- and ZO-based fine-tuning for Transformer models with the aim of providing theoretical limits on the model size that can be accommodated under a fixed on-chip memory budget. 

Throughout this section, as illustrated in Fig.~\ref{fig:param_llama}, we focus on a basic decoder-only Transformer architecture,   using the notation summarized in Table~\ref{tab:notations}. Specifically, we concentrate on traditional multi-head global self-attention and single-expert feedforward neural networks (FFN).  These choices exclude  the use of sliding-window attention \cite{beltagy2020longformer} and mixture-of-experts architectures \cite{shazeer2017, jacobs1991adaptive}. However, the analysis reported in this section can be generalized to these modern architectural variants by introducing variables such as window-size and expert count, without affecting the general conclusions. 


\begin{table}
    \centering
\caption{Notations used in text}
\label{tab:notations}
    \begin{tabular}{cc}
    \hline
         Symbol& Definition\\\hline\hline
         $N$& Context length\\
         $L$& Number  of layers\\
         $D$& Hidden dimension\\
         $H$& Number  of attention heads\\
 $K$& Key/ value heads \\
 $N_M$& Number  of MLPs in FFN layer\\
 $R$&MLP expansion factor (hidden  size $D \times R$)\\
 $V$&Vocabulary size\\
 $B$&Batch size\\
 $b$ & Bytes per parameter\\\hline
    \end{tabular}

\end{table}


\subsection{Parameter Count}
In order to establish the memory footprints of BP and MeZO, we start by counting the number of model parameters. For decoder-only LLMs, the model is largely composed of multi-head attention  blocks and FFN layers, along with embedding layers  and language model (LM) head. \\ \noindent $\bullet$ \textbf{Multi-head attention:} Given the embedding dimension $D$, the parameters per layer  encompass four projection matrices, typically denoted as $W_Q, W_K, W_V, W_O$, each of dimension $D\times D$. Specifically, each of the $H$ heads requires matrices of size $D \times D/H$. This yields a total  of   $4D^2$ parameters per layer. Memory-efficient variants include grouped query attention (GQA) \cite{ainslie-etal-2023-gqa}, in which $K<H$ query heads share a single key-value head, so that there are only $H/K$ key and value heads.   With GQA, the sizes of $W_K$ and $W_V$ thus reduce to $D\times D K/H$, bringing the total parameter count to $2D^2 (1+K/H)$. Note that setting $K=H$ recovers conventional multi-head attention.

 \noindent $\bullet$  \textbf{Feedforward neural networks:} Given an expansion factor $R$ between the embedding dimension and the FFN hidden dimension, the number of parameters per FFN layer is proportional to the product $DR$. In models such as LlaMa-2 and GPT-2, the value of $R$ is proportional to the hidden dimension $D$, e.g.,  $R \approx 2.67D$ for Llama-2 and $R=4D$ for GPT-2. Therefore, the total number of parameters is proportional to $D^2$, and the total parameter count for both models is approximated as $N_MLDR$. For  GPT-2 and LlaMa-2,  substituting the values of $\{N_M=2, R=4D\}$ and $\{N_M=3, R=2.67D\}$ respectively, we obtain the total parameter count for FFN layers as $8LD^2$ for both models  \cite{radford2019language}\cite{touvron2023llama}.

 \noindent $\bullet$   \textbf{Embedding layers, language model  head, and normalizing layers:} Both the embedding layer and the LM head have $VD$  parameters. This makes them bigger than projection matrices or MLPs in FFN. However, their contributions do not scale with the number of layers $L$, since they are not repeated at each layer.  Normalizing layers such as $\mathrm{RMSNorm}$ or $\mathrm{LayerNorm}$ have parameter counts proportional to $D$, so they only contribute a negligible fraction to the total parameter count.

Based on this analysis, we approximate the total parameter count as $2D^2(1+K/H) + 8D^2 + 2VD$. With conventional multi-head attention, i.e., when $K=H$, the parameter count becomes $12LD^2+2VD$.

\subsection{Activation Count }
With BP, activations must be cached for all layers.
As detailed in  \cite{korthikanti2023reducing}, and generalized here to GQA, the total memory of activation elements in bytes is approximately
\begin{equation}
A= BLND \left( 2 + 2b\bigg[7+\frac{K}{H}\bigg]
 + \frac{(2b+1)NH}{D} \right),
\label{eq:activations}
\end{equation} where $b $ is the number of bytes per parameter.

In contrast, MeZO does not require intermediate activation storage, as it only operates based on forward passes.

\subsection{Total Memory}
Assuming a stateless SGD optimizer, summing the contributions of model parameters and activations, the total memory requirement for BP can be estimated as\begin{equation}
    M_{\text{BP}} = 4bLD^2\bigg(5+\frac{K}{H}\bigg)+4bVD+A 
    \label{eq:mbpsgd}
\end{equation}
encompassing the contributions of model weights ($12bLD^2$), gradients ($12bLD^2$), the embedding layer, the LM head and corresponding gradients ($bVD$ each), and activations using \eqref{eq:activations}. With activation checkpointing, one can replace the number of layers $L$ in \eqref{eq:activations} with a smaller number, typically $\sqrt{L}$, as the remaining activations are recomputed as needed \cite{panchal2025avoidbp}.

Since MeZO stores no activations and does not involve gradient computation, the memory is in principle dominated by parameters.  In practice, however, the memory requirements depend on memory allocation and management policies, controlling the buffering of intermediate tensors \cite{huang2025reducing}. To account for this implementation-specific aspect, we introduce a factor $L'\leq L$, which dictates the maximum number of layers of activations stored by MeZO. This yields the memory requirement \begin{align}
M_{\text{MeZO}} &= 2bLD^2\bigg(5+\frac{K}{H}\bigg)+2bVD + \frac{L'}{L}A 
\label{eq:mmezo}
\end{align}

\subsection{Numerical Example}
Fig.~\ref{fig:BPvsMeZO} illustrates the memory requirements of BP, both with and without checkpointing (first and second row, respectively), and MeZO, along with the corresponding ratio between the memory requirements of BP and MeZO, when we vary context length $N$ (first column), number of layers $L$ (second column), and hidden dimension size $D$ (third column). We consider conventional multi-head attention, i.e., $K=H$.  For BP with activation checkpointing, following  \cite{panchal2025avoidbp}, we assume the  storage of activations for only $\sqrt{L}$ layers.  We adopt the LLaMa-2 7B model, for which the relevant variables in Table \ref{tab:notations} are set as $B=1$, $V=32000$, $N=2048$, $L=32$, $b=2$ (FP16), $H=32$, and $D=4096$. For the first column in Fig.~\ref{fig:BPvsMeZO}, the context length  $N$ was varied from 1 to 32768, with $D$ and $L$ fixed to their default values; while for the second and third columns the parameter $L$ was varied from 1 to 100 and $D$ from 512 to 32768, while keeping parameters $(N,D)$ and $(N,L)$ fixed  respectively. The dynamic memory allocation parameter $L'$ for MeZO was set to 1.

\noindent\textbf{Context scaling:} The top-left panel of Fig.~\ref{fig:BPvsMeZO} describes the memory requirements  with increasing context size for BP without checkpointing and for MeZO. 
The activation memory increase is quadratic with context, with BP memory requirement growth outpacing MeZO as per \eqref{eq:mbpsgd} and \eqref{eq:mmezo}.
As a result, the memory savings afforded by MeZO range from  2$\times$ at moderate contexts to up to 25$\times$ for $N=32768$.  As seen in the bottom-left part of Fig~\ref{fig:BPvsMeZO}. with activation checkpointing,  the advantage of MeZO over BP reduces to approximately 5$\times$ for long contexts, while remaining unchanged for smaller ones.

\noindent\textbf{Layer scaling:} Parameter memory scales linearly with $L$ for both BP and MeZO.  The top and bottom center panels in Fig.~\ref{fig:BPvsMeZO} show that for BP without checkpointing, the gain afforded by BP is smaller, around 1.2$\times$,  when  $L$ is small due to the contributions from the embedding layer and the LM head. These contributions diminish with larger $L$, an the memory savings  increases to 4.23$\times$ for larger $L$, with an asymptotic limit of 4.38. The effect of checkpointing on the MeZO/BP memory ratio is more significant for large layer sizes $L$. Specifically, the peak gain of around  2.28$\times$ at lower $L$ reduces slowly to 2.17$\times$ for $L=100$, approaching the asymptotic gain of 2$\times$.

\noindent\textbf{Hidden dimension scaling:} The dependence of parameter count with the hidden dimension $D$ is quadratic (see \eqref{eq:mbpsgd} and \eqref{eq:mmezo}). As shown in the right column of Fig.~\ref{fig:BPvsMeZO}, this  results in a rapid  increase in memory for both BP and  MeZO. For smaller $D$, the memory saving of MeZO can be as high as approximately 24$\times$, which reduces to 5$\times$ with BP checkpointing. With larger $D$, the memory saving ratio converges to 2.

Based on this  analysis,  we can conclude that, for the same memory budget, we can fine-tune a model that is at least 2$\times$ larger when using  MeZO as compared to using BP, with potentially several orders-of-magnitude larger savings for large contexts.

\section{Experiments}
\label{sec:experiments}

\begin{figure}
    \centering
    \includegraphics[width=0.9\linewidth]{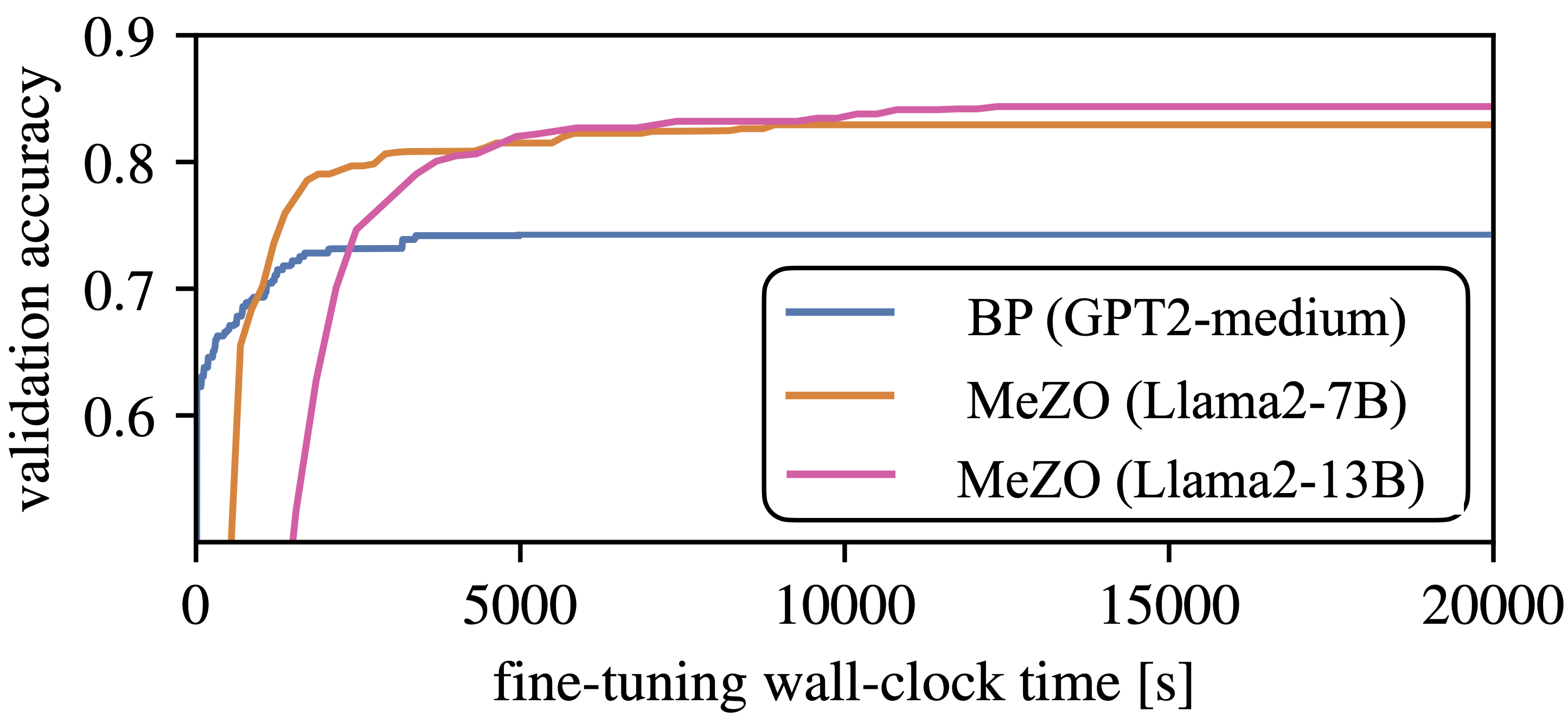}
    \caption{On-device fine-tuning for Boolq data set \cite{clark2019boolq} as a function of fine-tuning wall-clock time [s].  We consider MeZO with Llama2-7B and LlaMa2-13B, while BP adopts GPT2-medium model. The batch size is  $B=8$. All models  require a similar memory consumption of around 17 GB according to the analysis in Sec.~\ref{sec:memory} when setting $L'/L=0.15 $ for  Llama2-13B and the $L'/L=0.41$ for Llama2-7B.}
    \label{fig:BPvsMeZO_conv}
\end{figure}


This section validates the main claim of this work that MeZO can support the deployment of more powerful edge AI models when fine-tuning must be implemented on the device via numerical results. To this end, we adopt the BoolQ~\cite{clark2019boolq} and MultiRC~\cite{MultiRC2018} datasets with the prompting technique in~\cite{gao2020making} to support binary question answering. BoolQ consists of naturally occurring yes/no questions paired with supporting passages, while MultiRC extends this to a multi-sentence reasoning setting where each question may have multiple correct answers.

We consider GPT2-medium, Llama2-7B,  Llama2-13B, and Qwen3 models \cite{qwen3}, and we  include PEFT variants of BP and MeZO that fine-tune only a fraction  of the parameters~\cite{guo2025zerothorder}. All the experiments are carried out using a single H100 GPU. 

\subsection{Full Fine-Tuning}

We are interested in comparing models that have the same memory footprint. To this end, we leverage  the analysis in Sec.~\ref{sec:memory} assuming the maximum context length $N=1024$ with batch size $B=8$. Under these assumptions,  using (\ref{eq:mbpsgd}), fine-tuning GPT2-medium using BP has a memory requirement of around  17 GB. In contrast, the memory requirements of MeZO depend on the implementation-specific parameter $L'$ in (\ref{eq:mmezo}). In this experiment, we report results for specific choices of parameter $L'$ in order to provide insights into the impact of different memory management policies and hardware platforms, while the next subsection describes experimental results for the memory footprint. Recall that maximally efficient implementation corresponds to $L'/L=0$, while the BP memory requirement is recovered with $L'/L=1$. We evaluate two intermediate cases, namely $L'/L = 0.15$ and $L'/L = 0.41$. Using (\ref{eq:mmezo}), a memory of around 17GB is obtained with MeZO when  fine-tuning Llama2-7B if $L'/L = 0.41$ and Llama2-13B if $L'/L=0.15$.



In Fig.~\ref{fig:BPvsMeZO_conv}, we show the validation accuracy as a function of fine-tuning wall-clock time for the BoolQ dataset. We report the running maximum value for the accuracy.  We use vanilla SGD with a fixed learning rate for both BP and MeZO with the learning rate chosen by grid search. Specifically, we set the learning rate grid for BP as [5e-4, 5e-5, 5e-6] and for MeZO as [5e-7, 5e-8, 5e-9], and report the highest accuracy for both schemes. We fix the number of perturbations for MeZO to $5$ \cite{malladi2023fine}. Note that the number of perturbations does not affect the memory requirements of MeZO. 

Fig.~\ref{fig:BPvsMeZO_conv} confirms that MeZO has a slower fine-tuning convergence rate in terms of wall-clock time. However, thanks to the more capable model afforded by its more efficient memory usage, MeZO can eventually achieve a higher validation accuracy. For example, after about 2 hours, the accuracy of MeZO is around 82\%, while that of BP remains below 75\%.

\subsection{Parameter-Efficient Fine-Tuning}

We now consider PEFT, and evaluate memory requirements via GPU profiling,  instead of relying on the device-agnostic estimations derived in Sec. \ref{sec:memory}. We report the performance in terms of validation accuracy, as well as the measured GPU memory peak during training. We adopt  Qwen3-0.6B with BP and Qwen3-4B with MeZO, and include PEFT variants of BP and MeZO that fine-tune only a fraction $\rho=0.01$ of the parameters~\cite{guo2025zerothorder}. Specifically, the parameters to be fine-tuned are selected via a transferable static sparse mask, following the approach of \cite{guo2025zerothorder}. The coordinate-wise gradient squares are computed on a subset of WikiText~\cite{merity2017pointer} prior to fine-tuning, and the top $\rho = 0.01$ fraction of parameters are retained as the trainable subset. This mask is computed once on the pre-trained model and kept fixed throughout fine-tuning.

As shown in Table~\ref{tab:results}, MeZO on the 4B model still requires less memory than BP on the 0.6B model. The total memory of BP is reduced due the sparsity of the gradient update, though only modestly. For a sparsity level $\rho = 0.01$, i.e., only 1\% of the model parameters are updated, BP memory reduces only by approximately $12-13\%$. As discussed in Sec.~\ref{sec:memory}, activation memory dominates the total memory consumption, limiting the overall memory benefits of sparsity~\cite{tenison2026parameter}. From a performance perspective, both Sparse BP and Sparse MeZO achieve results comparable to their full-parameter counterparts. Overall, with MeZO a larger model can be fine-tuned compared to conventional and  sparse BP, even when gradient memory is reduced by around $99\%$, enabling better performance with the same memory budget. 


\section{Conclusions}
MeZO optimization eliminates the need to store activations and optimizer states, reducing training memory demands to the level of inference memory. This enables on-device learning with larger models and longer contexts. We have analyzed the relative memory requirements of MeZO compared to conventional BP-based fine-tuning and presented experimental results benchmarking their achievable accuracy on edge models. Overall, these findings position MeZO as a strong candidate for scenarios that require on-device fine-tuning, including agentic edge systems and neuromorphic architectures for continual learning. More research is required to specialize the main conclusions to neuromorphic systems and other models with dynamic sparsity at the level of activations. 

\begin{table}
    \centering
        \caption{Fine-tuning results for Qwen3 models on BoolQ~\cite{clark2019boolq} and MultiRC~\cite{MultiRC2018} datasets with a batch size of $B=4$}
    \label{tab:results}
    \setlength{\tabcolsep}{4pt}
    \begin{tabular}{lcccc}
        \toprule
        \textbf{Dataset} & \textbf{Model Size} & \textbf{Optimizer} & \textbf{Acc. (\%)} & \textbf{Memory (GB)} \\
        \midrule
        \multirow{4}{*}{BoolQ}
            & \multirow{2}{*}{0.6B} & BP        & 78.20 & 24.01 \\
            &                       & Sparse BP & 78.04 & 20.93 \\
            \cmidrule(lr){2-5}
            & \multirow{2}{*}{4B}   & MeZO        & 81.96 & 12.45 \\
            &                       & Sparse MeZO & 80.83 & 12.80 \\
        \midrule
        \multirow{4}{*}{MultiRC}
            & \multirow{2}{*}{0.6B} & BP        & 80.67 & 21.75 \\
            &                       & Sparse BP & 80.03 & 18.89 \\
            \cmidrule(lr){2-5}
            & \multirow{2}{*}{4B}   & MeZO        & 84.47 & 11.22 \\
            &                       & Sparse MeZO &  83.95   & 11.58 \\
        \bottomrule
    \end{tabular}
\end{table}

\bibliographystyle{IEEEtran}
\bibliography{refs}

@article{qwen3,
  title   = {Qwen Technical Report},
  author  = {Bai, Jinze and others},
  journal = {arXiv preprint arXiv:2309.16609},
  year    = {2023}
}

@inproceedings{
merity2017pointer,
title={Pointer Sentinel Mixture Models},
author={Stephen Merity and Caiming Xiong and James Bradbury and Richard Socher},
booktitle={International Conference on Learning Representations},
year={2017},
url={https://openreview.net/forum?id=Byj72udxe}
}

@article{liu2024sparse,
  title={Sparse mezo: Less parameters for better performance in zeroth-order llm fine-tuning},
  author={Liu, Yong and Zhu, Zirui and Gong, Chaoyu and Cheng, Minhao and Hsieh, Cho-Jui and You, Yang},
  journal={arXiv preprint arXiv:2402.15751},
  year={2024}
}

@inproceedings{
guo2025zerothorder,
title={Zeroth-Order Fine-Tuning of {LLM}s with Transferable Static Sparsity},
author={Wentao Guo and Jikai Long and Yimeng Zeng and Zirui Liu and Xinyu Yang and Yide Ran and Jacob R. Gardner and Osbert Bastani and Christopher De Sa and Xiaodong Yu and Beidi Chen and Zhaozhuo Xu},
booktitle={The Thirteenth International Conference on Learning Representations},
year={2025},
url={https://openreview.net/forum?id=myYzr50xBh}
}

@article{tenison2026parameter,
  title={Parameter Efficiency Is Not Memory Efficiency: Rethinking Fine-Tuning for On-Device LLM Adaptation},
  author={Tenison, Irene and Ahn, Stella and Kim, Miriam and Alshehri, Ebtisam and Kagal, Lalana},
  journal={arXiv preprint arXiv:2604.22783},
  year={2026}
}

@inproceedings{MultiRC2018,
    author = {Daniel Khashabi and Snigdha Chaturvedi and Michael Roth and Shyam Upadhyay and Dan Roth},
    title = {Looking Beyond the Surface:A Challenge Set for Reading Comprehension over Multiple Sentences},
    booktitle = {Proceedings of North American Chapter of the Association for Computational Linguistics (NAACL)},
    year = {2018}
}

@article{lialin2023scaling,
  title={Scaling down to scale up: A guide to parameter-efficient fine-tuning},
  author={Lialin, Vladislav and Deshpande, Vijeta and Rumshisky, Anna},
  journal={arXiv preprint arXiv:2303.15647},
  year={2023}
}

@article{chen2023fireact,
  title={Fireact: Toward language agent fine-tuning},
  author={Chen, Baian and Shu, Chang and Shareghi, Ehsan and Collier, Nigel and Narasimhan, Karthik and Yao, Shunyu},
  journal={arXiv preprint arXiv:2310.05915},
  year={2023}
}

@inproceedings{yang2024adazeta,
  title        = {{AdaZeta}: Adaptive Zeroth-Order Tensor-Train Adaptation for Memory-Efficient Large Language Model Fine-Tuning},
  author       = {Yang, Yifan and Zhen, Kai and Banijamali, Ershad and Mouchtaris, Athanasios and Zhang, Zheng},
  booktitle    = {Empirical Methods in Natural Language Processing (EMNLP)},
  year         = {2024},
  url          = {https://arxiv.org/abs/2406.08301}
}

@inproceedings{chen2025loho,
  title        = {Towards Efficient Low-Order Hybrid Optimizer for Language Model Fine-Tuning},
  author       = {Chen, Minping and Huang, You-Liang and Wen, Zeyi},
  booktitle    = {AAAI Conference on Artificial Intelligence},
  year         = {2025},
  note         = {Preprint},
  url          = {https://arxiv.org/abs/2409.18075}
}

@article{spall1998overview,
  title        = {An Overview of the Simultaneous Perturbation Method for Efficient Optimization},
  author       = {Spall, James C.},
  journal      = {Johns Hopkins APL Technical Digest},
  year         = {1998},
  volume       = {19},
  number       = {4},
  pages        = {482--492}
}

@misc{panchal2025avoidbp,
  title        = {The Cost of Avoiding Backpropagation},
  author       = {Panchal, Kunjal and others},
  year         = {2025},
  note         = {arXiv preprint},
  url          = {https://arxiv.org/abs/2506.21833}
}

@article{korthikanti2023reducing,
  title={Reducing activation recomputation in large transformer models},
  author={Korthikanti, Vijay Anand and Casper, Jared and Lym, Sangkug and McAfee, Lawrence and Andersch, Michael and Shoeybi, Mohammad and Catanzaro, Bryan},
  journal={Proceedings of Machine Learning and Systems},
  volume={5},
  pages={341--353},
  year={2023}
}

@article{malladi2023fine,
  title={Fine-tuning language models with just forward passes},
  author={Malladi, Sadhika and Gao, Tianyu and Nichani, Eshaan and Damian, Alex and Lee, Jason D and Chen, Danqi and Arora, Sanjeev},
  journal={Advances in Neural Information Processing Systems},
  volume={36},
  pages={53038--53075},
  year={2023}
}

@article{duchi2015optimal,
  title={Optimal rates for zero-order convex optimization: The power of two function evaluations},
  author={Duchi, John C and Jordan, Michael I and Wainwright, Martin J and Wibisono, Andre},
  journal={IEEE Transactions on Information Theory},
  volume={61},
  number={5},
  pages={2788--2806},
  year={2015},
  publisher={IEEE}
}

@article{touvron2023llama,
  title={Llama 2: Open foundation and fine-tuned chat models},
  author={Touvron, Hugo and Martin, Louis and Stone, Kevin and Albert, Peter and Almahairi, Amjad and Babaei, Yasmine and Bashlykov, Nikolay and Batra, Soumya and Bhargava, Prajjwal and Bhosale, Shruti and others},
  journal={arXiv preprint arXiv:2307.09288},
  year={2023}
}

@article{radford2019language,
  title={Language models are unsupervised multitask learners},
  author={Radford, Alec and Wu, Jeffrey and Child, Rewon and Luan, David and Amodei, Dario and Sutskever, Ilya and others},
  journal={OpenAI blog},
  volume={1},
  number={8},
  pages={9},
  year={2019}
}

@article{slamanig2025llms,
  title={From LLMs to Edge: Parameter-Efficient Fine-Tuning on Edge Devices},
  author={Slamanig, Georg and Corti, Francesco and Saukh, Olga},
  journal={arXiv preprint arXiv:2507.23536},
  year={2025}
}

@InProceedings{10.1007/978-3-031-99965-9_31,
author="Aralimatti, Rakshit
and Shakhadri, Syed Abdul Gaffar
and Kruthika, K. R.
and Angadi, Kartik Basavaraj",
editor="Arai, Kohei",
title="{Fine-Tuning Small Language Models for Domain-Specific AI: An Edge AI Perspective}",
booktitle="Intelligent Systems and Applications",
year="2025",
publisher="Springer Nature Switzerland",
pages="503--520",
isbn="978-3-031-99965-9"
}

@inproceedings{peng-etal-2024-pocketllm,
    title = "{P}ocket{LLM}: Enabling On-Device Fine-Tuning for Personalized {LLM}s",
    author = "Peng, Dan  and
      Fu, Zhihui  and
      Wang, Jun",
    booktitle = "Proceedings of the Fifth Workshop on Privacy in Natural Language Processing",
    month = aug,
    year = "2024",
    address = "Bangkok, Thailand",
    publisher = "Association for Computational Linguistics",
    url = "https://aclanthology.org/2024.privatenlp-1.10/",
    pages = "91--96"
}

@article{beltagy2020longformer,
  title={Longformer: The long-document transformer},
  author={Beltagy, Iz and Peters, Matthew E and Cohan, Arman},
  journal={arXiv preprint arXiv:2004.05150},
  year={2020}
}

@article{jacobs1991adaptive,
  title={Adaptive mixtures of local experts},
  author={Jacobs, Robert A and Jordan, Michael I and Nowlan, Steven J and Hinton, Geoffrey E},
  journal={Neural computation},
  volume={3},
  number={1},
  pages={79--87},
  year={1991},
  publisher={MIT Press}
}

@inproceedings{
shazeer2017,
title={ Outrageously Large Neural Networks: The Sparsely-Gated Mixture-of-Experts Layer},
author={Noam Shazeer and *Azalia Mirhoseini and *Krzysztof Maziarz and Andy Davis and Quoc Le and Geoffrey Hinton and Jeff Dean},
booktitle={International Conference on Learning Representations},
year={2017},
url={https://openreview.net/forum?id=B1ckMDqlg}
}

@article{10.1016/j.neucom.2023.127063,
author = {Su, Jianlin and Ahmed, Murtadha and Lu, Yu and Pan, Shengfeng and Bo, Wen and Liu, Yunfeng},
title = {RoFormer: Enhanced transformer with Rotary Position Embedding},
year = {2024},
issue_date = {Feb 2024},
publisher = {Elsevier Science Publishers B. V.},
address = {NLD},
volume = {568},
number = {C},
issn = {0925-2312},
url = {https://doi.org/10.1016/j.neucom.2023.127063},
doi = {10.1016/j.neucom.2023.127063},
journal = {Neurocomput.},
month = feb,
numpages = {12}
}

@inproceedings{clark2019boolq,
  title =     {BoolQ: Exploring the Surprising Difficulty of Natural Yes/No Questions},
  author =    {Clark, Christopher and Lee, Kenton and Chang, Ming-Wei and Kwiatkowski, Tom and Collins, Michael and Toutanova, Kristina},
  booktitle = {NAACL},
  year =      {2019}
}

@article{gao2020making,
  title={Making pre-trained language models better few-shot learners},
  author={Gao, Tianyu and Fisch, Adam and Chen, Danqi},
  journal={arXiv preprint arXiv:2012.15723},
  year={2020}
}

@article{huang2025reducing,
  title={Reducing GPU Memory Fragmentation via Spatio-Temporal Planning for Efficient Large-Scale Model Training},
  author={Huang, Zixiao and Hu, Junhao and Lin, Hao and Zhu, Chunyang and Tang, Yueran and Zhang, Quanlu and Guo, Zhen and Li, Zhenhua and Yan, Shengen and Zhu, Zhenhua and others},
  journal={arXiv preprint arXiv:2507.16274},
  year={2025}
}

@article{wang2026universally,
  title={Universally Empowering Zeroth-Order Optimization via Adaptive Layer-wise Sampling},
  author={Wang, Fei and Shen, Li and Ding, Liang and Xue, Chao and Liu, Ye and Ding, Changxing},
  journal={arXiv preprint arXiv:2604.18264},
  year={2026}
}

@inproceedings{ainslie-etal-2023-gqa,
    title = "{GQA}: Training Generalized Multi-Query Transformer Models from Multi-Head Checkpoints",
    author = "Ainslie, Joshua  and
      Lee-Thorp, James  and
      de Jong, Michiel  and
      Zemlyanskiy, Yury  and
      Lebron, Federico  and
      Sanghai, Sumit",
    editor = "Bouamor, Houda  and
      Pino, Juan  and
      Bali, Kalika",
    booktitle = "Proceedings of the 2023 Conference on Empirical Methods in Natural Language Processing",
    month = dec,
    year = "2023",
    address = "Singapore",
    publisher = "Association for Computational Linguistics",
    url = "https://aclanthology.org/2023.emnlp-main.298/",
    doi = "10.18653/v1/2023.emnlp-main.298",
    pages = "4895--4901"
}
\end{document}